\documentclass[conference]{IEEEtran}
\IEEEoverridecommandlockouts
\usepackage{cite}
\usepackage{amsmath,amssymb,amsfonts}
\usepackage{algorithmic}
\usepackage{graphicx}
\usepackage{textcomp}
\usepackage{xcolor}
\usepackage{here}
\usepackage{booktabs}
\usepackage{url}
\def\BibTeX{{\rm B\kern-.05em{\sc i\kern-.025em b}\kern-.08em
    T\kern-.1667em\lower.7ex\hbox{E}\kern-.125emX}}

\begin{document}

\title{University Building Recognition Dataset \\ for mission-oriented IoT sensor systems in Thailand}

\author{\IEEEauthorblockN{Takara Taniguchi}
\IEEEauthorblockA{\textit{The University of Tokyo} \\
hiroshi-tani@g.ecc.u-tokyo.ac.jp}
\and
\IEEEauthorblockN{Yudai Ueda}
\IEEEauthorblockA{\textit{The University of Tokyo} \\
ueda-yudai0913@g.ecc.u-tokyo.ac.jp}
\and
\IEEEauthorblockN{Atsuya Muramatsu}
\IEEEauthorblockA{\textit{The University of Tokyo} \\
muramatsu-atsuya870@g.ecc.u-tokyo.ac.jp}
\and
\IEEEauthorblockN{Kohki Hashimoto}
\IEEEauthorblockA{\textit{The University of Tokyo} \\
k-84mo10@g.ecc.u-tokyo.ac.jp}
\and
\IEEEauthorblockN{Ryo Yagi}
\IEEEauthorblockA{\textit{The University of Tokyo} \\
yagi-ryo143@g.ecc.u-tokyo.ac.jp}
\and
\IEEEauthorblockN{Hideya Ochiai}
\IEEEauthorblockA{\textit{The University of Tokyo} \\
ochiai@g.ecc.u-tokyo.ac.jp}
\and
\IEEEauthorblockN{Chaodit Aswakul}
\IEEEauthorblockA{\textit{Chulalongkorn University} \\
Chaodit.a@chula.ac.th}
}

\maketitle

\begin{abstract}
Many industrial sectors have been using of machine learning at inference mode on edge devices. 
Future directions show that training on edge devices is promising due to improvements in semiconductor performance. 
Wireless Ad Hoc Federated Learning (WAFL) has been proposed as a promising approach for collaborative learning with device-to-device communication among edges.
In particular, WAFL with Vision Transformer (WAFL-ViT) has been tested on image recognition tasks with the UTokyo Building Recognition Dataset (UTBR).
Since WAFL-ViT is a mission-oriented sensor system, it is essential to construct specific datasets by each mission. 
In our work, we have developed the Chulalongkorn University Building Recognition Dataset (CUBR), which is specialized for Chulalongkorn University as a case study in Thailand. 
Additionally, our results also demonstrate that training on WAFL scenarios achieves better accuracy than self-training scenarios.
Dataset is available in \url{https://github.com/jo2lxq/wafl/}.
\end{abstract}


\section{Introduction}
Machine learning on edge devices has become increasingly important because of the growing demand for distributed sensor systems.
In this context, federated learning is one of the approaches for machine learning on edge devices.
Centralized federated learning~\cite{LiTian2020,konečný2017federatedlearningstrategiesimproving} was originally proposed for machine learning on edges, which has been applied to many real-world problems, such as mobile keyboard prediction~\cite{hard2019federatedlearningmobilekeyboard}, COVID-19 warning systems~\cite{OUYANG2021124}, and financial risk management~\cite{pingulkar2024}.
Meanwhile, decentralized federated learning~\cite{ochiai2022wirelessadhocfederated, ochiai2023} has also been proposed to tackle problems such as privacy issues and the single point of failure that are attributed to traditional federated learning.

In particular, Wireless Ad hoc Federated Learning (WAFL)~\cite{ochiai2022wirelessadhocfederated} has been proposed as a framework for decentralized federated learning.
WAFL has been applied to a generative adversarial network~\cite {tomiyama2023} and the robustness of model poisoning attacks~\cite{tezuka2022}, respectively.
WAFL-ViT~\cite{ochiai2023} is proposed as a collaborative training method for Vision Transformer~\cite{dosovitskiy2021imageworth16x16wordsViT} for the mission-oriented image recognition task with the UTokyo Building Recognition Dataset (UTBR).
While sensor systems such as WAFL are expected to be constructed for a specific mission, such as an image recognition task of UTBR, few datasets have been investigated for WAFL-ViT.

In this study, we consider the image recognition task for the smart-campus application in Chulalongkorn University.
Therefore, we create the Chulalongkorn University Building Recognition Dataset (CUBR) under the assumption of an IoT sensor system.
CUBR compounds of 32 buildings, where the number of images of each building is about 100.
To consider an IoT environment of Chulalongkorn University, we assume that 10 devices are distributed in the campus of Chulalongkorn University.
Moreover, we evaluated the dataset by using collaborative learning.
In our evaluation, we compare Vision Transformer (ViT)~\cite{dosovitskiy2021imageworth16x16wordsViT} with VGG~\cite{simonyan2014very}, Resnet~\cite{he2016deep}, and Mobilenet~\cite{howard2017mobilenets} with WAFL and self-training cases, respectively.
As a result of the experiment, we show that WAFL enhances the accuracy compared to self-training cases.

Our contributions are summarized as follows:
\begin{itemize}
    \item[-] We tackle evaluating the decentralized federated learning model WAFL.
    \item[-] We develop the new dataset CUBR, which is composed of 32 labeled buildings for the image recognition task.
    \item[-] We demonstrate that WAFL works for the created CUBR compared to the self-training cases.
\end{itemize}

\begin{figure*}[htbp]
    \centering
    \includegraphics[width=1\linewidth]{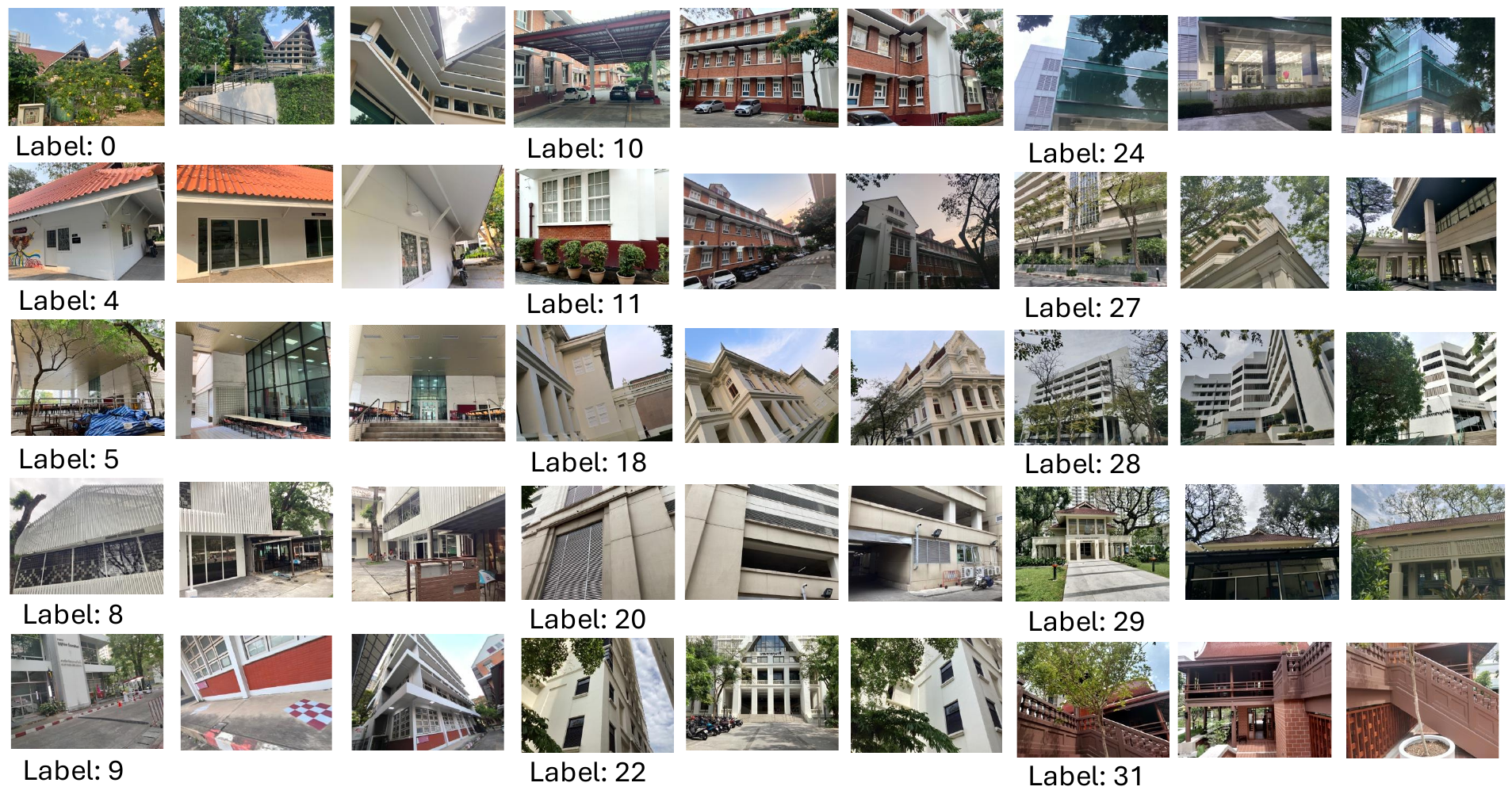}
    \caption{Examples of CUBR, where images are taken from aside, afar, and various angles. Even within the same building, there are different features according to the conditions in which images are taken.}
    \label{fig:examples}
\end{figure*}

\section{Wireless Ad Hoc Federated Learning}
In this section, we explain related works about WAFL and its dataset.
Federated learning has been proposed as a method to enhance model generalization by performing machine learning across multiple edge devices~\cite{pmlr-v54-mcmahan17a}.
In recent years, applications of federated learning in the real world have been proposed.
Hard \textit{et al.}~\cite{hard2019federatedlearningmobilekeyboard} suggested the application of federated learning in predictive models for mobile keyboards, while Ouyang \textit{et al.}~\cite{OUYANG2021124} applied federated learning to the COVID-19 early warning system.
In addition, Pingulkar \textit{et al.}~\cite{pingulkar2024} investigated the architecture of asset management using federated learning.

However, traditional federated learning models~\cite{LiTian2020,konečný2017federatedlearningstrategiesimproving} are centralized federated learning, which contain a single point of failure.
To tackle the problem of a single point of failure, Ochiai \textit{et al.}~\cite{ochiai2022wirelessadhocfederated} proposed Wireless Ad hoc Federated Learning (WAFL), which is a fully decentralized and mission-oriented federated learning model under the idea of the work by Frodigh \textit{et al.}~\cite{Johansson2000}. 
WAFL applications have been investigated for several missions since WAFL is a mission-oriented sensor system.
While Tezuka \textit{et al.}~\cite{tezuka2022} investigated the robustness of WAFL for the model poissoning attacks, Tomiyama \textit{et al.}~\cite{tomiyama2023} addressed distributed generative adversarial networks using WAFL.

In this context, Ochiai \textit{et al.}~\cite{ochiai2023} proposed WAFL-ViT~\cite{ochiai2023}, which is a collaborative training method for Vision Transformer~\cite{dosovitskiy2021imageworth16x16wordsViT}.
In the following, we briefly explain the foundation of WAFL-ViT. We denote $n$ as devices participating in the training. 
Let ${\rm adj}(n)$ and $W^n$ be the set of neighbor nodes of device $n$ and parameters of MLP heads of device $n$, where $\lambda$ means a hyperparameter between 0 to 1, respectively. 
Then, the algorithm of parameter exchange is given as follows:
\begin{align}
W^{n} \leftarrow W^{n}+\lambda\frac{\sum_{k\in {\rm adj}(n)}(W^{k}-W^{n})}{\vert {\rm adj}(n)\vert +1}
\end{align}
Using this algorithm, the parameters of their MLP heads are updated.
In training, parameters of MLP heads are fine-tuned with their \textit{local} images once per round.
To avoid overfitting to the \textit{local} images, all parameters of each ViT in each device are not fine-tuned.

To evaluate WAFL-ViT, Ochiai \textit{et al.}~\cite{ochiai2023} also created the UTokyo Building Recognition Dataset (UTBR) composed of 10 labeled buildings, which is assumed to be utilized for a smart-campus service. 
While creating a dataset of the mission-oriented task for WAFL is essential, there are few datasets for the image recognition task, except for UTBR. 
Hence, we tackle creating a large dataset of Chulalongkorn University for the image recognition task.

\begin{figure}[htbp]
    \centering
    \includegraphics[width=0.8\linewidth]{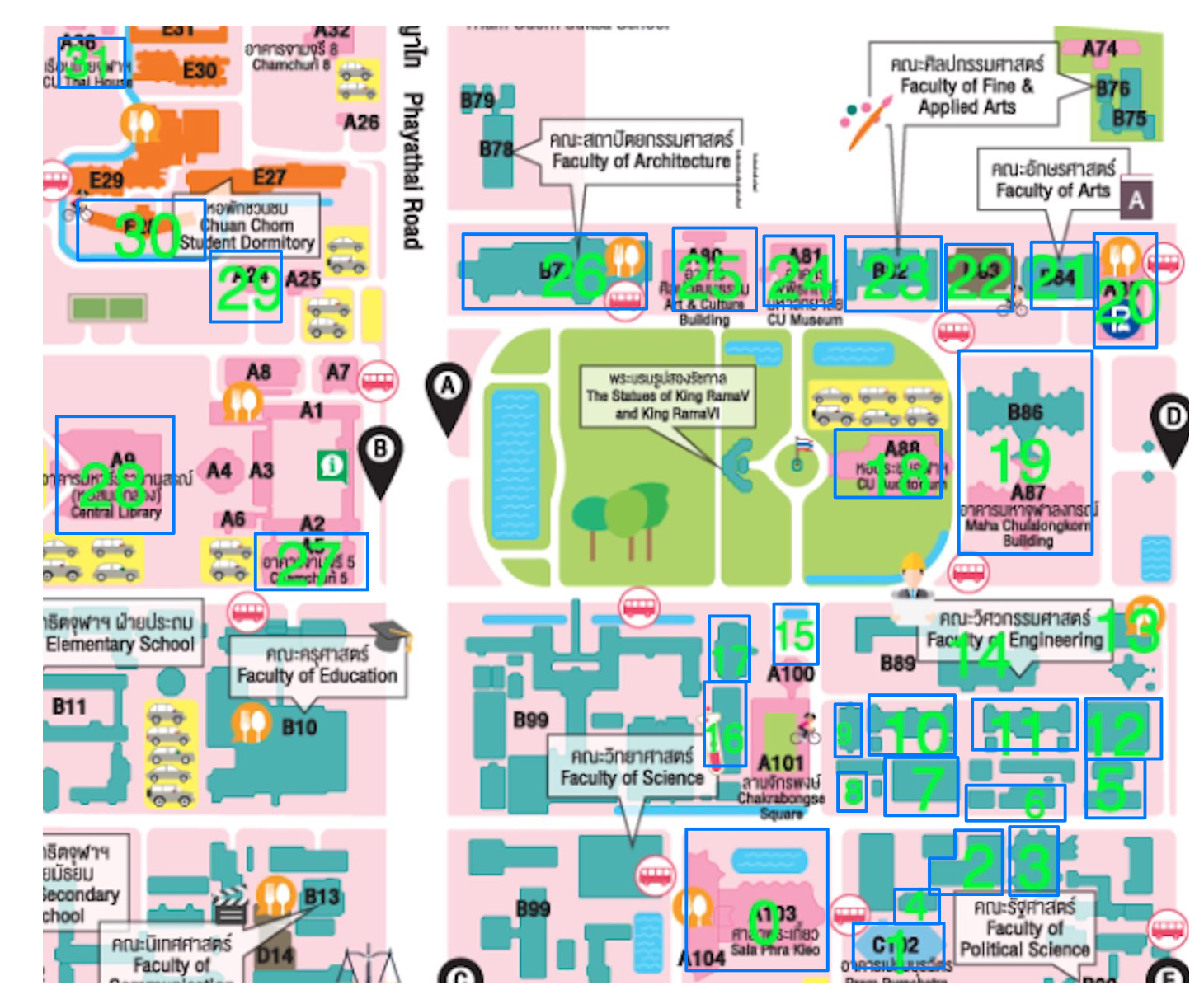}
    \caption{Target area where the photos were taken and correspondence between the label number and each building.}
    \label{fig:maps}
\end{figure}

\section{Chulalongkorn University Building Recognition Dataset}
In this section, we explain the contents of the created dataset in detail.
For WAFL, which is a mission-oriented sensor system, we created a Chulalongkorn University Building Recognition Dataset (CUBR).
The correspondence between each building and the label number in the map is given in Fig.~\ref{fig:maps}, and the number of photos of labeled buildings is shown in Table~\ref{tbl:trainval}, respectively.
As shown in Table~\ref{tbl:trainval}, 32 distinct buildings within Chulalongkorn University were selected as targets for this dataset. 
Photographs were captured using 4 different devices.
This multi-device approach was adopted to introduce variability in image quality, resolution, and sensor characteristics, which is crucial for developing robust recognition models. 

The selection criteria prioritized buildings with varying architectural designs, sizes, and functional purposes to ensure a diverse set of visual features. 
Similar to the UTBR dataset~\cite{ochiai2023}, these photos were captured from a wide variety of environments and conditions. 
For example, images were taken from different distances, various viewing angles, and under diverse illumination conditions. 
Fig.~\ref{fig:examples} illustrates some examples of these varied perspectives and conditions. 
Furthermore, we intentionally included images with partial occlusions and cluttered backgrounds to reflect real-world complexities. 
The data collection was primarily conducted over a period of a month.

\begin{table}[htbp]
\centering
\caption{The number of photos for training and validation.}\label{tbl:trainval}
\begin{tabular}{c@{\hspace{3pt}}c@{\hspace{3pt}}c@{\hspace{3pt}}c@{\hspace{3pt}}c@{\hspace{3pt}}c@{\hspace{3pt}}c@{\hspace{3pt}}c@{\hspace{3pt}}c@{\hspace{3pt}}c@{\hspace{3pt}}c@{\hspace{3pt}}c@{\hspace{3pt}}c@{\hspace{3pt}}c@{\hspace{3pt}}c@{\hspace{3pt}}c@{\hspace{3pt}}c@{\hspace{3pt}}}
\hline
\textbf{Label} & 0  & 1   & 2  & 3  & 4  & 5  & 6   & 7   & 8   & 9  & 10  & 11 & 12  & 13 & 14  & 15 \\
\hline
\textbf{Train} & 98 & 125 & 93 & 88 & 83 & 93 & 103 & 101 & 104 & 88 & 120 & 86 & 104 & 84 & 107 & 110  \\
\hline
\textbf{Val} & 25 & 32  & 26 & 21 & 20 & 24 & 26  & 26  & 26  & 22 & 31  & 22 & 26  & 22 & 22  & 28 \\
\hline
\end{tabular}

\vspace{0.2cm}

\begin{tabular}{c@{\hspace{3pt}}c@{\hspace{3pt}}c@{\hspace{3pt}}c@{\hspace{3pt}}c@{\hspace{3pt}}c@{\hspace{3pt}}c@{\hspace{3pt}}c@{\hspace{3pt}}c@{\hspace{3pt}}c@{\hspace{3pt}}c@{\hspace{3pt}}c@{\hspace{3pt}}c@{\hspace{3pt}}c@{\hspace{3pt}}c@{\hspace{3pt}}c@{\hspace{3pt}}c@{\hspace{3pt}}}
\hline
\textbf{Label} & 16  & 17 & 18  & 19  & 20 & 21 & 22  & 23  & 24  & 25 & 26 & 27  & 28  & 29 & 30 & 31 \\
\hline
\textbf{Train} & 103 & 84 & 120 & 126 & 99 & 84 & 107 & 104 & 120 & 92 & 88 & 103 & 102 & 83 & 86 & 106 \\
\hline
\textbf{Val} & 26  & 22 & 30  & 32  & 25 & 22 & 27  & 25  & 30  & 23 & 28 & 26  & 26  & 21 & 22 & 27 \\
\hline
\end{tabular}
\end{table}

\begin{table}[htbp]
\centering
\caption{Accuracy overview – higher better. \\
WAFL-ViT achieved the best in the IID case.}
\begin{tabular}{@{}lrr@{}}
\toprule
Model          & Accuracy & Standard Deviation \\ \midrule
WAFL-ViT       & 0.861                        & 0.007                                  \\
WAFL-Resnet    & 0.833                        & 0.022                                  \\
WAFL-VGG       & 0.732                        & 0.008                                  \\
WAFL-Mobilenet & 0.791                        & 0.015                                  \\
SELF-ViT       & 0.619                        & 0.017                                  \\
SELF-Resnet    & 0.596                        & 0.025                                  \\
SELF-VGG       & 0.498                        & 0.023                                  \\ 
SELF-Mobilenet & 0.561                        & 0.018    \\               \bottomrule
\end{tabular}
\end{table}

\section{Experiment}

\subsection{Implementation detail}
We evaluated the performance with the CUBR dataset.
In our experiment, we compare Vision Transformer-B/16 with ResNet-152, VGG-19-BN, and MobileNet-V2 with WAFL and self-training cases, as well as previous research by Ochiai \textit{et al.}~\cite{ochiai2023}.
The description of "WAFL-*" and "SELF-*" follows from Ochiai \textit{et al.}~\cite{ochiai2023} as well.
For "WAFL-*", we have carried out training up to 500 epochs for self-training and up to 1500 epochs for collaborative training.
For "SELF-*", we have carried out training up to 2000 epochs for self-training.

We apply random waypoint mobility (RWP)~\cite{bettstetter2002} to the simulation for the movement of each device.
We assume that all images are uniformly distributed to each device, which is denoted as \textit{IID} in \cite{ochiai2023}.

\subsection{Result}
Table~\ref{tbl:trainval} shows the accuracy and the standard deviation of each model.
The accuracy is calculated at 10 nodes by test data.
Mean and standard deviation are calculated by $ \text{(the number of labels)}  \times \text{(the number of nodes)} = 32 \times 10 = 320$ accuracies.
WAFL-ViT achieved the best accuracy among all models.
Also, the result shows that the performance of all models is improved from self-training cases to WAFL cases.

\begin{figure}[htbp]
    \centering
    \includegraphics[width=0.8\linewidth]{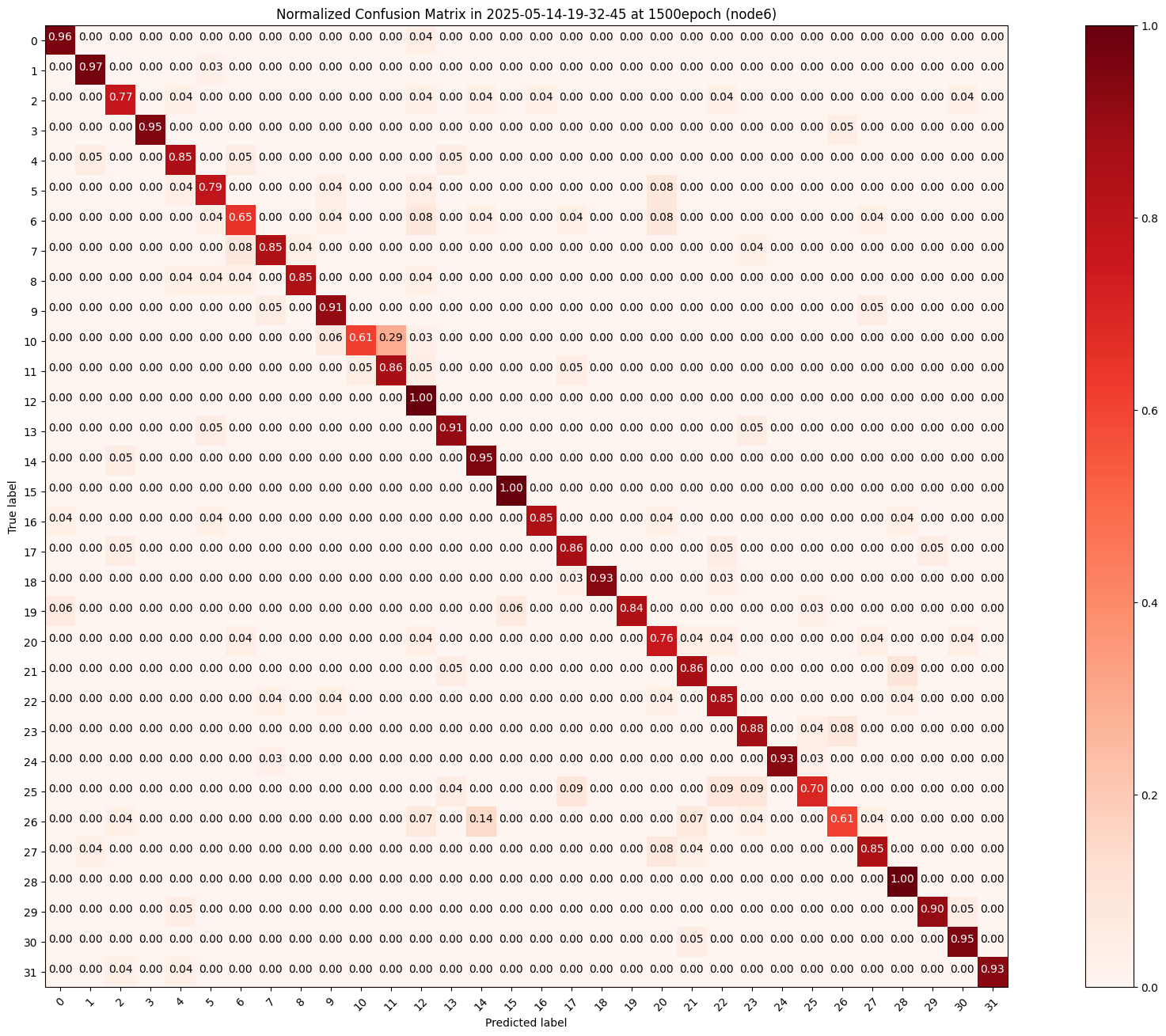}
    \caption{Confusion Matrix of WAFL-ViT at epoch 2000 (device 6)}
    \label{fig:confusion_matrix}
\end{figure}

The WAFL models generally outperform the SELF models in terms of accuracy, with WAFL-ViT being the most accurate and consistent. 
SELF-VGG exhibits the least favorable performance, both in accuracy and consistency. This comparative analysis highlights the superiority of WAFL architectures over SELF architectures across different neural network structures.
Fig.~\ref{fig:confusion_matrix} shows the confusion matrix of node 6.
We obtain that distinguishing between the label 10 and label 11 is difficult.

\section{Conclusion}
In this study, we considered the decentralized federated learning model WAFL-ViT for the image recognition task in Chulalongkorn University.
Since WAFL-ViT is a mission-oriented model, we created the Chulalongkorn University Building Recognition Dataset (CUBR) for WAFL-ViT, which is composed of 32 buildings.
We carried out the simulation with an assumption of 10 distributed devices and obtained the result that WAFL-* models achieved better accuracy than SELF-* models, respectively.

\newpage

\bibliography{bibtex/bib/IEEEabrv,bibtex/bib/IEEEexample}{}

\begin{thebibliography}{00}
\bibitem{b1} G. Eason, B. Noble, and I. N. Sneddon, ``On certain integrals of Lipschitz-Hankel type involving products of Bessel functions,'' Phil. Trans. Roy. Soc. London, vol. A247, pp. 529--551, April 1955.
\bibitem{b2} J. Clerk Maxwell, A Treatise on Electricity and Magnetism, 3rd ed., vol. 2. Oxford: Clarendon, 1892, pp.68--73.
\bibitem{b3} I. S. Jacobs and C. P. Bean, ``Fine particles, thin films and exchange anisotropy,'' in Magnetism, vol. III, G. T. Rado and H. Suhl, Eds. New York: Academic, 1963, pp. 271--350.
\bibitem{b4} K. Elissa, ``Title of paper if known,'' unpublished.
\bibitem{b5} R. Nicole, ``Title of paper with only first word capitalized,'' J. Name Stand. Abbrev., in press.
\bibitem{b6} Y. Yorozu, M. Hirano, K. Oka, and Y. Tagawa, ``Electron spectroscopy studies on magneto-optical media and plastic substrate interface,'' IEEE Transl. J. Magn. Japan, vol. 2, pp. 740--741, August 1987 [Digests 9th Annual Conf. Magnetics Japan, p. 301, 1982].
\bibitem{b7} M. Young, The Technical Writer's Handbook. Mill Valley, CA: University Science, 1989.
\end{thebibliography}
\bibliographystyle{ieeetr}

\end{document}